# Context based Analysis of Lexical Semantics for Hindi Language

Mohd Zeeshan Ansari and Lubna Khan

Department of Computer Engineering, Jamia Millia Islamia, New Delhi, India.

mzansari@jmi.ac.in, lubna.java@gmail.com

**Abstract.** A word having multiple senses in a text introduces the lexical semantic task to find out which particular sense is appropriate for the given context. One such task is Word sense disambiguation which refers to the identification of the most appropriate meaning of the polysemous word in a given context using computational algorithms. The language processing research in Hindi, the official language of India, and other Indian languages is restricted by unavailability of the standard corpus. For Hindi word sense disambiguation also, the large corpus is not available. In this work, we prepared the text containing new senses of certain words leading to the enrichment of the sense-tagged Hindi corpus of sixty polysemous words. Furthermore, we analyzed two novel lexical associations for Hindi word sense disambiguation based on the contextual features of the polysemous word. The evaluation of these methods is carried out over learning algorithms and favorable results are achieved.

**Keywords:** Natural Language Processing, Lexical Semantics, Word Sense Disambiguation.

## 1    Introduction

All natural languages have certain words with multiple meanings called homonyms. The automatic selection of the appropriate meaning of such words is a challenging task in natural language processing. Humans can easily comprehend the appropriate meaning of such word using the context in which it is used. However, the relationship between the meaning and the context is not well understood by machines because the computational representation of context is considerably complex. When a particular word has different meanings, also called senses, pertaining to the contexts in which it is used, it is called polysemous. This characteristic produces a great deal of complexity in the processing of natural languages. In lexical semantics of natural languages, the context is closely related to the specific task, domain and underlying language. Since certain words or phrases that occur in a given text may be interpreted in multiple ways, the significance of context is to determine its appropriate sense. Therefore, for the au-tomatic identification of the particular sense of a word, the context analysis is necessary. Consequently, the Word Sense Disambiguation (WSD) task is to determine the most appropriate sense of a word in a given context. Most of the WSD techniques consider context as the text surrounding the polysemous word, usually in a fixed size window keeping the word in the middle. Several approaches to solve WSD task for English and European languages are present in literature. Most of them are classified under three major approaches: Knowledge-based, supervised and unsupervised approaches [1,2,7,8,13,15,24]. Enough WSD research based on semi-supervised and hybrid approaches is carried out, with English being the primary language.

The studies on complex language processing tasks such as machine translation, information extraction, question answering, sentiment analysis, etc. involving Indian languages, especially Hindi, the official language of India, is constrained by unavailability of large standard corpora. As WSD is involved in many such tasks, it is therefore, quiet challenging for Indian languages as they are rich in morphology. Also, the development of several resources like WordNet, machine-readable dictionaries, language corpora, etc. are still under progress [32,35]. In this article, we present the work on Hindi written in Devanagari script, which is the official language of India. We explore the Hindi WSD task by the interpretation and analysis of the context in a variety of ways. The Sense Tagged Hindi Corpus, used in this work was developed under the Technology Development for the Indian Languages (TDIL) project, Government of India [30]. This corpus is available for research on Hindi Word Sense Disambiguation which consist of polysemous words and their instances, see Table 1. We enriched this corpus by the inclusion of more senses in the case of two existing words. This enriched sense tagged Hindi corpus is used to investigate lexical and semantic attributes significantly to Hindi WSD task. Our contribution to Hindi WSD is three folds (1) propose additional sense of two words (i) 'बाल' meaning 'भुट्टे का बाल' (corn silk), and (ii) 'कदम' meaning 'कदम का पेड़' (bur flower) and their instances to the existing corpus (2) we explore two novel attribute associations for Hindi WSD and test them on a range of window size, (3) present comparative analysis of their performance with respect to the methods found in literature. For a comparative evaluation of our methods, we also constructed the attributes defined by Singh et. al. [29].

In this work, we investigate various attribute associations for Hindi WSD task. The feature vector is constructed using the associations of local context, collocation, bag of words after stop word removal and *vibhakti*. These vectors of features when tested with the different window size of text produce significant results. The overall work is organized in the sections as follows. Section 2 shows related work on general WSD task, highlighting the challenges faced by researchers in WSD. Section 3 introduces Hindi WSD tasks. Section 4 gives the basic definitions of attributes of WSD task. Section 5 gives the detailed description of our work, i.e., the suggested attribute associations, the existing methods with illustrative examples. Section 6 gives detailed description of the dataset preparation, experimental setup and the analysis and discussion of results. Section 7 presents the conclusion and future scope work.

Table 1. Corpus statistics.

|  | Original Corpus | Enriched Corpus |
|---|---|---|
| Number of words | 381875 | 383008 |
| Number of instances | 7506 | 7570 |
| Number of polysemous words | 60 | 60 |

## 2  Related Word

In lexical semantics, the word sense disambiguation principally involves two key tasks, firstly the determination of potential senses (or meanings) of each word, and secondly, the classification of each word to its probable sense with low error and computational complexity. The challenges identified for the WSD task are (1) discreteness of the senses, (2) differences between dictionaries, (3) amount of samples and semantic knowledge available. The discreteness of the senses can be divided into coarse-grained and fine-grained levels. A human can easily comprehend coarse-grained which deals with homographs, but it is quite difficult for a human to understand fine-grained level. The number of samples and semantic knowledge available can be enhanced by building them manually which involves large costs. Numerous approaches and algorithms are built for disambiguating words based on a given context. In initial phases, the disambiguation methods were typically evaluated on a small dataset consisting of words with scarce and strong sense differences, e.g. Yarowsky [4] reported 96% precision for twelve words with only two strong sense distinctions. Subsequently, large number of approaches were examined and huge effort was dedicated to solve this task. ACL SIGLEX Workshop concluded with an open challange "Tagging Text with Lexical Semantics: Why, What and How? WSD is perhaps the great open problem at the lexical level of NLP" [7,8,10]. Similarly, the outcomes of the Senseval, Senseval-2 and Senseval-3 in which systems submitted in these conferences did not achieve eighty percent accuracy on lexical sample as well as all words tasks [14,16,23].

Bayesian classification models applied to several investigations in WSD are considerably successful [22,31]. Due to its simplicity, the Naive Bayes algorithm is the preferred choice to obtain the preliminary results on supervised WSD [12,18,21]. Mooney [6] considered seven supervised learning algorithms namely Naive Bayes, decision tree, k-nearest neighbor, perceptron, slogic-based conjunctive and disjunctive normal form learners and a decision list learner. Subsequently, they were tested on the ambiguous word 'line' having six senses and a comparison is drawn among them, which showed favorable results on Naive Bayes classifier and perceptron. The words surrounding the ambiguous word were used as features for the classifiers. Leacock et. al. [3] used Naive Bayes classifier and merged topic and local context to achieve comparable accuracy for the combination in order to disambiguate among nouns, a verbs and an adjectives. Pederson [12] took an ensemble of several simple Naïve Bayes classifiers to improve WSD accuracy using nine different window sizes of left and right of context: 0-5,10,20 and 50. A measure of 88% and 89% accuracy was achieved on the two datasets. Le and Shimazu [19] studied Naive Bayes classifier for performing WSD task utilizing enriched features. They combined features represented by sequence of words in a local context and collocations using features derived from a forward sequential selection algorithm. The accuracy obtained was 92.3% for four common test words and 72.7% for nouns and 66.4% for verbs when tested on large DSO corpus. A decomposable model proposed by Bruce and Wiebe [11] is complex model which considered dependence graph of various dependent and independent features. The primary weakness of this approach is estimation of large number of parameters, which are related to the number of different mix of the interdependent characteristics. Therefore, this technique required a large size of dataset as well as computational cost to significantly estimate the parameters. In contrast to this, Pederson and Rebecca [9] proposed an unsupervised method on unlabeled text for the identification of an optimal model, with high performance and low cost in parameter estimation. Zhong and Ng developed [27] a system based on a supervised learning approach, a flexible framework that achieved good results on several sensEval and semEval tasks. Lee et. al. [20] investigated support vector machines on multiple knowledge sources to obtain the results for a lexical task and reported that their method performed fairly well. Chaplot, et. al. [33] developed an unsupervised model using Maximum A Posteriori Inference Query built on a Markov Random Field using WordNet and Standford Parser. It is a graphical model which was evaluated on English all word dataset and showed better and fast results compared to the existing best unsupervised models [25]. The entropy maximization approach estimates probabilities on the basis of making a small but necessary number of assumptions. This property is satisfied only by the probability distribution with the highest entropy. The advantage of such models is that even poor features may be applied accurately, therefore, such models are also found relevant in case of WSD [5,17].

Singh, et. al. [29] have put forward the initial efforts on Hindi WSD task using supervised classification, principally by the application of Naive Bayes algorithm in sense tagged Hindi corpus. They considered eleven feature vectors based on local context, collocations, bag of words and *vibhakti*. The precision obtained from their model is 77.52% for bag of words, and 86.11% for nouns words in feature vector after applying morphology. A precision of 56.49%, by including *vibhakti* in the feature vector is also reported. The task for Hindi WSD supervised approach was further extended using sense annotated Hindi corpus, dictionary definitions and semantic relationships to assign weightage to words present within the context of polysemous words . The context of the polysemous word is a sequence of words appearing in window with the polysemous word at the center. For evaluation, sense tagged corpus consisting of sixty Hindi words (nouns) using sense definitions and semantic relations obtained from Hindi WordNet, were utilized. The results reported show that overall average precision and recall values were 78.98% and 73.41% respectively.

**Table 2.** Excerpts of Hindi text containing the polysemous word 'हार' in two senses.

| Sense | |
|---|---|
| Sense 1 | Paragraph1. निर्माता-निर्देशक करण जौहर ने ट्विटर पर लिखा है केकेआर की **हार** से बहुत दुख हुआ। खेल को खेल भावना से देखना चाहिए और **हार** स्वीकार करनी चाहिए। |
| Sense 2 | Paragraph 2. न्यूयॉर्क। हीरे का **हार** पहनी एक बार्बी गुड़िया न्यूयॉर्क में रेकॉर्ड कीमत में नीलाम हुई है। अपनी तरह की ये अकेली बार्बी डॉल काला लिबास पहने हुई है और उसके गले में एक कैरेट का चौकोर गुलाबी हीरे का **हार** है। ये गुड़िया में बनाया गया था और तबसे लेकर आज तक इसका रूप कई बार बदला है। सबसे बड़ी नीलामी का रेकॉर्ड बनाने वाली बार्बी गुड़िया को ऑस्ट्रेलिया के एक गहनों के डिजायनर स्टीफानो कैन्टुरी ने बनाया है। |

Hadni et. al. [34] reported WSD task for Arabic text categorization using feature selection methods implemented on SVM and NB classifier and achieved a precision of 72% and 71% respectively. Naseer and Sharmad reported the work on NB classifier for Urdu WSD. They selected four ambiguous Urdu words (1 noun and 3 verbs) having high frequencies appearing in the corpus used with varying window sizes of $\pm 3$, $\pm 5$, $\pm 7$ around the target word. The maximum accuracy achieved was 98.33% for window size 7x7 when tested on three verbs and one noun. They observed that an increase in window size resulted in a enhanced performance [26].

## 3 Word Sense Disambiguation for Hindi Language

The application of Word Sense Disambiguation on polysemous words present in Hindi language text written in Devanagari script is called Hindi Word Sense Disambiguation. A Hindi polysemous word does also have a different meaning in different contexts. The text excerpts in Table 2. is the text in the Hindi language which illustrates one such word 'हार'. It can be clearly observed that there are two senses of the word 'हार', in the first paragraph it means 'पराजय' (defeat) and in the second paragraph, it means 'माला' (necklace). The more details of senses of word 'हार' is presented in Table 3.

**Table 3.** Senses of 'हार'

| Sense | Synonyms | Meaning | Use in sentence |
|---|---|---|---|
| Sense 1 | पराजय, आपजय (defeat) विघात, असफलता (failure) मात (checkmate) अभिभव (disgrace) | पराजित होने की अवस्था या भाव (defeated state or expressions) | इस चुनाव में उसकी हार निश्चित है (in this election his defeat is certain) |
| Sense 2 | माला (garland) नेकलेस (necklace) अवतंस,अवतन्स (garland) आभूषण (jewelry) | गले में पहनने का एक प्रकार का सोने, चाँदी आदि का गहना (a type of gold or silver jewelry used for wearing around the neck) | उसने हीरे का हार पहन रखा है (she is wearing a diamond necklace) |

## 4 Lexical Attributes for Hindi WSD task

The basic lexical attributes used in sense disambiguation are local context, collocation, bag of words, bag of words after stop words removal and *vibhakti* which are defined below. Illustrative examples[#] of each method are given throughout this section.

**Definition 1.** Local context ($l_j$) is defined as the collection of words surrounding the ambiguous word in a given piece of text with a window size of j. The local context feature set denoted by $l_2$, contain words of local context in a window size ±2, i.e. two words from the left and two words from the right of the ambiguous word. The example may be given as $l_2$ = ['हीरे', 'का', 'हार', 'पहनी', 'एक'][#].

**Definition 2.** Collocation ($c_j$) is defined as the group of those sequence of words that include the target word. In a window size of j, the collocation feature set, $c_j$, is the group of those sequences of size 2 to j+1 words which contain the ambiguous word. The collocation feature set denoted by $c_2$, contain collocation in a window size ±2, i.e. word sequence of size 2 and 3 which include the ambiguous word. The example may be given as $c_2$ = ['हीरे का हार ', 'का हार', 'हार पहनी', 'हार पहनी एक', 'का हार पहनी'][#].

**Definition 3.** Bag of words ($b_j$) is the simple bag of j words in the left and right of the ambiguous words in a window size ±j, but without removing the stop words from the text. The example of window size ±2 may be given as $b_2$ = ['हीरे', 'का', 'हार', 'पहनी', 'एक'][#].

**Definition 4.** Bag of words after stop word removal ($b^*_j$) is the bag of j words in the left and right of the ambiguous word in a window size ±j after removing the stop words from the text. The example may be given as $b^*_2$ = ['', 'हीरे ', 'हार', 'पहनी'][#].

**Definition 5.** *Vibhakti* ($v_j$) are set of words considered important constituent of Hindi grammar, referring to the relationship between verbs and other constituents typically nouns or pronouns in a sentence. The *vibhakti* involved in this work are, $v$ = [ ने(ne) , को(ko) , से(se) , के(ke) , लिए(liye) , का(ka) , की(ki) , में(mein) , पर(par) , हे(hey) , अरे(arey)] [27]. The feature set is defined by making bag of words with only *vibhakti* in the left and the right of the target word with a window size ±j. The example feature set may be given as, $v_2$ = ['', 'का', '',''][#].

[#]Table 2. Paragraph 2.

**Table 4.** Dataset description for Hindi WSD task.

| Number of senses | Hindi Polysemous words |
|---|---|
| 2 | अशोक, कांड, कोटा, क्रिया, गल्ला, गुना, गुरु, ग्राम, घटना, चंदा, चारा, जीना, जेठ, डब्बा, डाक, , सोना, हल, ढाल, तान, ताव, तिल, तीर, तुलसी, दक्ष, दर, दाद, दाम, धन, धुन, माँग, लाल, विधि, शेर, सीमा, हार |
| 3 | अंग , अंश, अचल, उत्तर, कमान, कुंभ, क्रांटर, खान, चरण, तेल, थान, फल, बाल, मत, मात्रा, वचन, वर्ग, संक्रमण, संबंध |
| 4 | कदम, कलम, धारा, मूल |
| 5 | चाल, टीका |

## 5  Lexical Attribute Associations for Hindi WSD task

In order to construct a novel set of lexical attribute associations, we defined lexical association feature set on a range of window size. This set is obtained by a unique combination of basic attributes, viz. local context, collocation, bag of words after stop words removal and *vibhakti*. We proposed two novel attribute associations and tested them on window size range from ±2 to ±5. These are defined as follows

**Collocation and bag of words after stop words removal ([c+b*]j).** It is the combination of local context and bag of words after stop words removal of window size j. The feature set $[c+b^*]_j$ is obtained by merging the basic features of collocation $c_j$ and bag of words after stop word removal $b^*_j$ of window size ±j. Therefore, the feature set $[c+b^*]_2$ is defined as the combination of $c_2$ and $b^*_2$ of window size j=±2. Using the same example text as used in previous examples, the feature set of $[c+b^*]_2$ is obtained as ['हीरे का हार', 'का हार', 'हार पहनी', 'हार पहनी एक', 'का हार पहनी', 'हीरे', 'हार', 'पहनी']#. Similarly, the feature sets of window size ±3, ±4, ±5 are also obtained.

**Local context, collocation and vibhakti ([l+c+v]j).** It is the combination of three basic features, local context, collocation and *vibhakti*. The feature $[l+c+v]_j$ is obtained by merging the basic features of local context $l_j$, collocation $c_j$ and *vibhakti* $v_j$ of window size ±j. Therefore, the feature $[l+c+v]_2$ is defined as the combination of feature $l_2$, $c_2$ and $v_2$ of window size j=±2. Using the same example text as used in previous examples, the feature set of $[l+c+v]_2$ is obtained as ['हीरे', 'का', 'हार', 'पहनी', 'एक', 'हीरे का हार ', 'का हार', 'हार पहनी', 'हार पहनी एक', 'का हार पहनी', '', 'का', '', '']#. Similarly, the feature sets of different window size of ±3, ±4, ±5 are also constructed.

**Table 5.** Precision, Recall and F-Score

| Method | Window | P | R | F |
|---|---|---|---|---|
| Collocation + bow | 5 | **0.80** | **0.85** | **0.82** |
| Collocation + local context + vibhakti | | 0.74 | 0.80 | 0.77 |
| Singh et. al. | | 0.77 | 0.82 | 0.79 |
| Bow after stop word removed | | 0.76 | 0.81 | 0.78 |
| Collocation + bow | 4 | 0.79 | 0.84 | 0.82 |
| Collocation + local context + vibhakti | | 0.74 | 0.80 | 0.77 |
| Singh et. al. | | 0.76 | 0.82 | 0.79 |
| Bow after stop word removed | | 0.75 | 0.81 | 0.78 |
| Collocation + bow | 3 | 0.78 | 0.83 | 0.80 |
| Collocation + local context + vibhakti | | 0.74 | 0.80 | 0.77 |
| Singh et. al. | | 0.76 | 0.81 | 0.78 |
| Bow after stop word removed | | 0.74 | 0.80 | 0.77 |
| Collocation + bow | 2 | 0.77 | 0.82 | 0.79 |
| Collocation + local context + vibhakti | | 0.76 | 0.81 | 0.78 |
| Singh et. al. | | 0.76 | 0.81 | 0.78 |
| Bow after stop word removed | | 0.70 | 0.76 | 0.73 |

The attribute association suggested by Singh et. al. (2015) [29] for Hindi Word Sense Disambiguation is *local context with collocation* ($[l+c]_j$). It is a combination of local context and collocation of window size j. This set is obtained by merging the basic feature sets $l_j$ and $c_j$ as defined at the beginning of this section. The feature set $[l+c]_2$ consists of words in local context and collocation in window size j=±2, which can also be obtained by merging the basic features $l_2$ and $c_2$. The example may be given as $[l+c]_2$ = ['हीरे', 'का', 'हार', 'पहनी', 'एक', 'हीरे का हार', 'का हार', 'हार पहनी', 'हार पहनी एक', 'का हार पहनी'] [#]. This feature combination $[l+c]_j$ was analyzed for different window size having a range from j = ±5 to ±25 [27].

## 6 Analysis of Lexical Relations for Hindi WSD

### 6.1 Experimental Setup

The dataset is generated using the sense tagged Hindi corpus, also used by Singh et. al. [29] which consists of sixty polysemous Hindi nouns with their senses. Table 4. presents the complete set of all such words along with the number of senses for each word. To this corpus, we added one more sense of two words, 'बाल' and 'कदम'. It was identified that the word 'बाल' has one more sense 'भुट्टे का बाल' (corn silk) which is not present in the corpus. Similarly, it was also identified word 'कदम' has one more sense

'कदम का पेड़' (bur flower) which is also not present in the corpus. Subsequently, the instances of both words were collected and merged with the corpus leading to the final dataset having additional 64 instances of 1133 words more, than original corpus.

Although the corpus is organized in a defined format, we further preprocess it for our experimental settings. Finally, we divide the dataset into training and testing data according to ratio 3:1. We performed a series of experiments using the proposed as well as previously existing attribute associations and also using the basic attributes alone. All the methods are examined on the window size of range from ±2 to ±5.

### 6.2 Analysis of results

By the application of the Naïve Bayes classification algorithm, we predicted the appropriate sense of each ambiguous word. We predicted it for each of the test samples, and reported the sense accuracy as the *number of correct sense predictions* divided by *total number of test sample*. We computed the metrics such as precision, recall and F1-measure for the two proposed attribute associations as well as methods of Singh et. al. [29] which are presented in Table 5. It is observed from the analysis that the proposed attribute associations for collocation and bag of words after stop word removal with window size ±5, i.e. $[c+b*]_5$, performs best and achieves the highest precision, recall and accuracy values of 0.80, 0.85 and 0.85 respectively. Moreover, this combination method with window size in the range from ±3 to ±5 outperforms the other methods. We deduce that this combination is best suited for our problem and on increasing the window size greater than ±5 much higher performance can be achieved. The second proposed combination of local context, collocation and *vibhakti* shows poor performance than other methods because of the introduction of *vibhakti*. The reason being that there is no sense related words in it. This can be endorsed by the observation that it alone shows the worst performance. Therefore, we deduce that the *vibhakti* do not make any contribution in the performance improvement of the model when used alone as well as when used in combination, moreover it degrades the performance when used in combination. The performance of basic attributes without combination show reduced performance as compared to the proposed and existing attribute associations. The bag of words after stop words removal of window size ±5, i.e. $b*_5$ is the best among the basic attributes. It is, therefore, considered most appropriate for the comparison in the analysis throughout, although, the results of local context and collocation have also been obtained from the experiments.

## 7  Conclusion

The present work explores the lexical semantics for the Hindi language in Devanagari script. A contribution is made by the addition one new sense each in case of two polysemous words which contains 1133 Hindi Devanagari words under 64 instances. The investigation is carried out to examine the effect of several lexical attribute associations over Hindi sense disambiguation. Realizing that the neighbouring words in the context of the ambiguous word play a vital role in feature vector formation, two novel attribute associations are proposed. These methods are successfully tested along with the existing and basic methods on the corpus of 60 polysemous words. The precision, recall and F1- measure are computed for each feature vector. One of our proposed combination method performs best as compared to rest of the methods which is observed from the figures of outcomes obtained from the experiments. The second proposed method validates the fact that *vibhakti* do not make any contribution in disambiguation of senses. A scope of examining the proposed methods on a higher range of window size is left for future.